# Flexible Decomposition Algorithms for Weakly Coupled Markov Decision Problems

**Ronald Parr**
Computer Science Department
Stanford University
Stanford, CA 94305-9010
parr@cs.stanford.edu

## Abstract

This paper presents two new approaches to decomposing and solving large Markov decision problems (MDPs), a partial decoupling method and a complete decoupling method. In these approaches, a large, stochastic decision problem is divided into smaller pieces. The first approach builds a cache of policies for each part of the problem independently, and then combines the pieces in a separate, light-weight step. A second approach also divides the problem into smaller pieces, but information is communicated between the different problem pieces, allowing intelligent decisions to be made about which piece requires the most attention. Both approaches can be used to find optimal policies or approximately optimal policies with provable bounds. These algorithms also provide a framework for the efficient transfer of knowledge across problems that share similar structure.

## 1 Introduction

The Markov Decision Problem (MDP) framework provides a formal framework for modeling a large variety of stochastic, sequential decision problems. It is a well-understood framework with well-known on-line and off-line algorithms for determining optimal behavior (see e.g. Puterman (1994)). The limitations of this framework are also well-known: compliance with the Markov property generally requires a very fine granularity description of the environment. Thus, most interesting problems have large state spaces.

One of the main research thrusts for MDPs has been the development of methods for large state spaces. A major complicating factor in this line of research is the apparent non-decomposability of MDPs — the utility or value of any state can, in general, be affected indirectly by the cost structure and the dynamics of any other state. This thwarts efforts to decompose MDPs into completely independent subproblems and complicates efforts to reduce computation time through parallelization.

While some progress has been made on understanding some very special cases where MDPs may be decomposed into independent subproblems (Singh, 1992; Lin, 1997), much of the effort has focused on methods that decompose MDPs into "communicating" subproblems (Bertsekas & Tsitsiklis, 1989; Dean & Lin, 1995). In these iterative methods, information about subproblem solutions is communicated to neighboring subproblems. The solution for each subproblem may need to be updated many times until a globally optimal solution is obtained.

This paper considers a special, but fairly general class of problem decompositions where each subproblem is "weakly" coupled with the neighboring subproblems. This means that the number of states connecting the two subproblems is small, a relationship that appears naturally in many problems. For example, the problem of moving from one's office to one's house has this structure: one's office is a small region that is connected by a much smaller region, the door, to an external corridor. Many other offices may be connected to this corridor, each with a similar structure. The corridor could be fairly large and connected to other corridors by relatively small intersection regions. Most buildings have a small number of doorways that connect them to the streets outside. Each street has a relatively small number of points where it connects to other streets. One such street connects to the house one calls home, which is itself an aggregation of weakly connected pieces. An MDP is weakly coupled if it can be divided into two or more subproblems that are weakly coupled with each other. Figure 1 shows a simple navigation MDP divided into four rooms, each of which can be considered a subproblem.[1]

This paper extends the communicating MDP solution methods with the goal of avoiding iterating between different subproblem solutions by constructing a cache of policies independently for each subproblem that is guaranteed

---

[1] Similar examples and pictures are used by Precup and Sutton (1998) and Hauskrecht, Meuleau, Boutilier, Kaelbling, and Dean (1998).

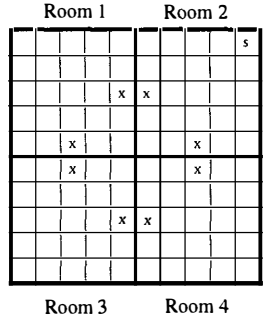

Figure 1: weakly coupled MDP. There is a reward in room 2, indicated with a $. Out-space states are identified with an X.

*a priori* to provide performance within a constant of the optimal, regardless of the structure of the other subproblems. This permits a complete decoupling of the MDP into independent subproblems that can be solved in parallel and then recombined in a single step. The decoupling is achieved through two new algorithms that build a cache of policies iteratively by discovering the point at which the current cache performs the worst, then adding a new policy to the cache to cover the worst case.

The efficient manipulation of policy caches also provides a formal basis for the transfer of knowledge across problems with similar substructures. The simplest case of this occurs when the reward structure for a problem changes. Suppose, for example that the reward in the navigation problem is moved from room 2 to room 4. Policy caches devised for rooms 1 and 3 can be can be transferred to the new problem. Similarly, if one's destination is now a cafe instead of home, the policies designed for one's office and the containing building should transfer to the new problem.

Since the number of possible policies for a subproblem is exponential in the number of states in the subproblem, there may exist problems and accuracy requirements for which the size of the policy cache will be exponential. In these cases there still will be some benefit to constructing a small policy cache, even if it does not provide the desired accuracy guarantees. This paper presents an algorithm that augments standard communicating MDP algorithms with the use of a policy cache. The policy cache can be used to determine lower *and* upper bounds on the values that states in the subproblem can assume, and this provides a means of deciding when it is worth using a cached solution and when it is worth producing a new subproblem solution. This is particularly useful in determining if subproblem solutions from a related problem can be applied to a new one.

## 2   Markov Decision Problems

To review the basic MDP framework, an MDP is a 4-tuple, $(\mathcal{S}, \mathcal{A}, \mathcal{T}, \mathcal{R})$ where $\mathcal{S}$ is a set of *states*, $\mathcal{A}$ is a set of *ac-*

*tions*, $\mathcal{T}$ is a *transition model* mapping $\mathcal{S} \times \mathcal{A} \times \mathcal{S}$ into probabilities in $[0, 1]$, and $\mathcal{R}$ is a *reward function* mapping $\mathcal{S} \times \mathcal{A}$ into real-valued rewards. Algorithms for solving MDPs can return a *policy*, $\pi$, that maps from $\mathcal{S}$ to $\mathcal{A}$, or a real-valued *value function* $V$. In this paper, the focus is on infinite-horizon MDPs with a discount factor $\beta$. The aim in these problems is to find an optimal policy $\pi^*$, that maximizes the expected discounted total reward of the agent, or to find an approximately optimal policy that comes within some bound of optimal.

Value iteration, policy iteration or linear programming can be used to determine the optimal policy for an MDP. These algorithms all use some form of the Bellman equation (Bellman, 1957):

$$V^*(s) = \max_a R(s, a) + \beta \sum_{s'} \mathcal{T}(s, a, s') V^*(s').$$

When the Bellman equation is satisfied, the maximizing action for each state is the optimal action.

For a particular policy, the Bellman equation becomes a system of linear equations:

$$V_{\pi_1}(s) = R(s, \pi_1(s), s') + \beta \sum_{s'} \mathcal{T}(s, \pi_1(s), s') V_{\pi_1}(s').$$

These can be solved to determine, $V_{\pi_1}$, the value of following $\pi_1$ from any state. The *Bellman error* for a particular policy at a particular state is the difference between the value function for that policy and the right-hand side of the Bellman equation:

$$BE(V_\pi(s)) = \max_a R(s, a) + \beta \sum_{s'} \mathcal{T}(s, a, s') V_\pi(s') - V_\pi(s)$$

For any policy, the maximum Bellman error over all states, $BE(V_\pi) = \max_s BE(V_\pi(s))$, is a well-known bound on the distance from the optimal value function (Williams & Baird, 1993):

$$V^*(s) \le V_\pi(s) + \frac{BE(V_\pi)}{1 - \beta}$$

This is used below to produce error bounds on policy caches.

## 3   A Class of Decomposition Algorithms

The first step for a decomposition method for MDPs is the division of the state space into disjoint subsets, $G_1 \ldots G_m$. In the simple navigation problem of Figure 1, each room could be a subset. For each subset, $G_i$, there exists a set of states not in $G_i$ that are reachable in one step from $G_i$. Call this the *out-space* of $G_i$. In Figure 1, the out-space of the top-left room contains one state in the room to the right, and one in the room below. The *in-space* of region $G$ is defined as the set of states inside of $G$ reachable in one step from a region outside of G.

The next step is the introduction of a set of policy caches, $\Pi_1 \ldots \Pi_m$, defined over each of the regions. It is well known that the optimal assignment of policies to regions can be determined by solving a "high-level" reduced decision problem defined over only the states in the out-spaces of the regions. This reduced decision problem removes all but the out-space states from the problem. Actions in the reduced problem correspond to assignments of policies to regions in the original decision problem. This transformation is the basic insight of Forestier and Varaiya (1978) and it follows as a special case of the hierarchical results in Parr and Russell (1998). The approach is also investigated in (Hauskrecht et al., 1998). This type of problem also can be viewed as Semi-Markov decision problem (SMDP), where each low-level policy becomes a primitive SMDP action as in Parr (1998).

In Figure 1, the high level problem would contain just the eight specially marked states. An action in the high level problem would correspond to a decision to adopt some policy from the cache upon entering a room, and staying with this policy until the next out-space state is reached. The solution to the high-level problem may produce a *non-stationary* policy at the low-level, which means that the actions taken in any room may depend upon the manner in which the room is entered. A non-stationary policy of this type can be converted easily to a stationary policy that is at least as good (See Parr (1998)).

The relationship between the size of the out-spaces and the complexity of the high-level problem should make the importance of weak coupling clear. If the size of the out-spaces approaches the size of the original MDP, then the high-level decision problem will be as difficult as the original MDP.

An algorithm that *completely* decomposed an MDP would produce a $\Pi_i$ for each $G_i$, combine these to produce an optimal or approximately optimal overall solution, and never need to revise any of the $\Pi_i$. Unless the $\Pi_i$ are chosen very carefully, or the caches are very large, combinations of policies in the initial policy caches may not suffice. There are several approaches to revising the policy caches. One extreme end of this spectrum is the approach in Sutton, Precup, and Singh (1998), where policies and low-level actions are mixed together in the same SMDP. This sacrifices the reduction in computational complexity obtained from solving a reduced decision problem in favor of a guarantee of obtaining optimality. Another approach considered by Dean and Lin (1995) updates each $\Pi_i$ directly. Dean and Lin considered a special case in which the old policies were discarded at each iteration, and a new policy was computed for each region based upon the high-level decision problem's current estimates for the value of the out-space states. Note that the high-level decision problem was actually a trivial value determination problem and not really a decision problem since $|\Pi_i| = 1$ for all $i$.

The approach advocated by Dean and Lin is guaranteed to converge to the optimal policy. However, it is just one special case of a general class of methods that must converge. Any reasonable scheme that improves the policies in the regions and propagates those improvements through the high-level decision problem is guaranteed to produce an optimal policy as long as no regions "starve", i.e., never have their policies improved. This result follows directly from the observation that the high-level problem of assigning policies to rooms is really just an SMDP where the set of permitted actions for the SMDP are just the set of possible policies defined over regions.

The algorithms in this paper all aim to minimize the number of policies that are computed for MDP subproblems. The extent to which this can be minimized is a measure of how effectively an MDP has been decomposed. If each subproblem requires only a small cache of candidate solutions, this means that the subproblem solutions are relatively independent. These are precisely the situations in which a large computational benefit is reaped from decomposition, since the MDP can be divided and conquered by solving a reasonable number of small subproblems. The size of the policy caches also gives some measure of the parallelizability of the problem. If a region can be solved with a small cache of policies, this suggests that the entire cache could be constructed *a priori* as a completely independent subprocess.

The following section describes several algorithms for constructing policy caches with minimal knowledge of how the subproblem is connected to overall MDP. These algorithms aim to minimize the size of the cache, while ensuring that solutions using the cache will be within a bound of optimal. The succeeding section describes a scheme for working with policy caches for which optimality bounds have not been established *a priori*. This method efficiently establishes bounds on the benefit of adding a new policy to a cache, based upon the current contents of the cache.

## 4    Complete decoupling

This section presents algorithms that find a policy cache, $\Pi$, for a particular region, $G$, such that $\Pi$ is guaranteed to provide policies that are within a constant of optimal when a high level problem using $\Pi$ for $G$ is solved. The only assumptions that are made about the regions to which $G$ connects is that the states assume values on $[V_{\min} \ldots V_{\max}]$.

Define $\mathcal{V}_G^{\mathcal{O}}$, as a vector of values that the states in the out-space of $G$ can take on (the subscript will be dropped when there can be no confusion about the region in question). The *fan-out* of a region is defined as the dimension of this vector.

In addition to storing a cache of policies it is useful to store a cache of functions, $f_{\pi_i}(s, \mathcal{V}^{\mathcal{O}})$ for each $\pi_i \in \Pi$. Each $f_{\pi_i}(s, \mathcal{V}^{\mathcal{O}})$ is linear function that provides the value of any state $s \in G$ as a linear function of $\mathcal{V}^{\mathcal{O}}$. For any policy, $\pi_i$, it is quite straightforward to determine these functions by solving a system of linear equations, as in Parr (1998).

The goal in constructing a policy cache for a region is to produce a cache such that for every possible value of the corresponding out-space states, there is a policy in the region's cache for which the performance in the region will be within a bound of optimal. A policy, $\pi$, for region, $G$, is optimal with respect to $\mathcal{V}^{\mathcal{O}}$ if $\pi$ is the solution to the MDP defined just over the states in $G$, with the assumption that states in the out-space of $G$ are absorbing states with values locked at the value of the corresponding entry in $\mathcal{V}^{\mathcal{O}}$. In room 1 of the four-room example, the optimal policy for $\mathcal{V}^{\mathcal{O}}$ would be determined by solving an MDP with just the states in room 1 and the two connecting states in room 2 and room 3. The value of the connecting state in room 2 would be treated as a constant with value $\mathcal{V}^{\mathcal{O}}[0]$ and the value of the connecting state in room 3 would be a constant with value $\mathcal{V}^{\mathcal{O}}[1]$. A policy, $\pi$, is said to be $\epsilon$-optimal with respect to $\mathcal{V}^{\mathcal{O}}$ if $\forall s \in G$, $BE(V_\pi) \le \epsilon$ when the values of the states in the out-space of $G$ are fixed by $\mathcal{V}^{\mathcal{O}}$.

For any state and any value of $\mathcal{V}^{\mathcal{O}}$, there must be one policy in the cache that appears at least as good as all of the others. A policy, $\pi$ *dominates at* $t$ for a particular $\mathcal{V}^{\mathcal{O}}$, if $f_\pi(t, \mathcal{V}^{\mathcal{O}}) \ge f_{\pi_j}(t, \mathcal{V}^{\mathcal{O}})$ $\forall j$. This means that the low-level policy, $\pi$, appears to be the best high level action at state $t$ for a particular $\mathcal{V}^{\mathcal{O}}$. A cache of policies is $\epsilon$-*optimal at* $t$ if, for any $\mathcal{V}^{\mathcal{O}}$, the dominating policy is $\epsilon$-optimal. A cache of policies is $\epsilon$-optimal if it is $\epsilon$-optimal at all $t$ in the in-space of $G$.

**Theorem 1** *If an MDP is divided into $m$ regions, $G_1 \ldots G_m$ and an $\epsilon$-optimal cache of policies, $\Pi_1 \ldots \Pi_m$, is determined for each region, these policies can be combined to produce a globally $\frac{\epsilon}{(1-\beta)}$-optimal policy by solving an MDP with at most $\sum_i |\mathcal{V}^{\mathcal{O}}_{G_i}|$ states and $\sum_i |\Pi_i|$ actions.*

**Proof:** As in Section 3 a high-level decision problem can be defined over the states $U = \bigcup_i G_i$. The solution to this decision problem assigns values to the states in $U$ and assigns dominating policies to each state in the in-space of each region in $U$ based upon these values. Consider an arbitrary region $G$ and the policy, $\pi$, assigned to it when it is entered at state $t$. By the definition of $\epsilon$-optimality, $BE(V_\pi(s)) \le \epsilon$ for all $s$ in $G$. Since this will be true for all states in all regions, the maximum Bellman error for any policy that will be used in any region will be less than $\epsilon$, which means that the maximum Bellman error for the entire problem will be less than $\epsilon$, which is sufficient to ensure $\frac{\epsilon}{(1-\beta)}$-optimality.∎

This theorem provides a means of combining a set of approximately optimal solutions to produce a global solution that is also approximately optimal. Of course, if $\epsilon = 0$, then the solution will be optimal. The following subsections will describe three algorithms for constructing $\epsilon$-optimal policy caches.

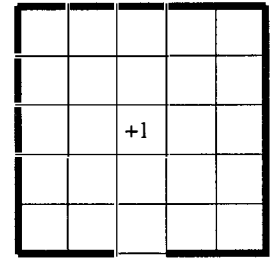

Figure 2: A simple MDP with a one state out-space.

## 4.1 The $\epsilon$-grid Approach

The most straightforward approach to devising a policy cache, as described in Hauskrecht et al. (1998), is to create an $\epsilon$ resolution grid over the space of possible values for $\mathcal{V}^{\mathcal{O}}$, then produce an optimal policy with respect to each grid point. This can be proved to be sufficient by observing that the value of any policy can change by at most $\epsilon$ when moving within one cell of the grid. This bounds the loss from using a the policy defined for the nearest grid point instead of the optimal policy. This result is established formally using an alternative argument in (Hasukrecht, 1998).

The main problem with the $\epsilon$-grid approach is that it can require a huge number of policies, $(\frac{V_{max}-V_{min}}{\epsilon})^d$. This will be unmanageable unless the range of values is very small, the fan out of the region is very small, or $\epsilon$ is very large.

## 4.2 Value Space Search

This section presents an algorithm that aims to avoid constructing an exponential number of policies by searching through $\mathcal{V}^{\mathcal{O}}$ space to find a point at which the current policy cache is not adequate. If such a point is found, a new policy is added to the cache, and the process is repeated until no points can be found for which the current cache is inadequate. The following formal results are the basis of the value space search algorithm:

**Lemma 1** *If the dominating policy in a policy cache is used for any $\mathcal{V}^{\mathcal{O}}$, then the value function for any state, $s$, is a piecewise-linear, convex function of $\mathcal{V}^{\mathcal{O}}$.*

**Proof:** This follows from the observation that using the best policy means taking the maximum over a set of linear policy functions. ∎

This lemma is demonstrated with a simple example using the model in Figure 2, where there is a room with just one exit, but a $+1$ absorbing reward state in the center of the room. The value function for this room can be displayed in two dimensions since $\mathcal{V}^{\mathcal{O}}$ is just a scalar. Figure 3 shows linear functions for the value of the top-left state for the optimal policy for $\mathcal{V}^{\mathcal{O}} = 0$, $\mathcal{V}^{\mathcal{O}} = 1.09$, and $\mathcal{V}^{\mathcal{O}} = 2$.

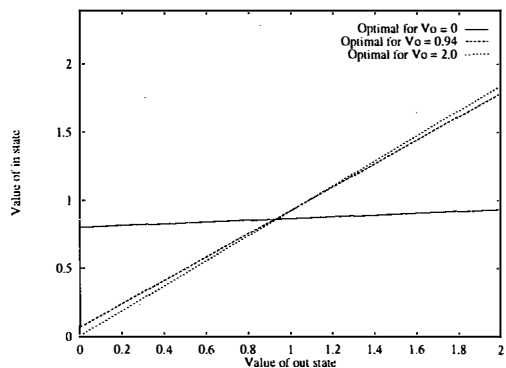

Figure 3: Two policies, and an upper surface bounding the their distance from the optimality.

**Theorem 2** *For region $G$, with $n$ states, $a$ actions, and policy cache, $\Pi = \pi_1 \ldots \pi_m$, the point in $\mathcal{V}^{\mathcal{O}}$ space for which the Bellman error of the dominating policy is largest, can be determined in time that is polynomial in $n$, $a$, $m$, and $d$.*

**Proof:** This is achieved by means of a linear program. For all $t$ in the in-space of $G$, for all $s$ in $G$, for all $a$, and for all $\pi \in \Pi$, the following linear program is solved:

*Maximize:*

$$R(s, a) + \sum_{s'} T(s, a, s') f_\pi(s', \mathcal{V}^{\mathcal{O}}) - f_\pi(s, \mathcal{V}^{\mathcal{O}})$$

*Subject to:*

$$f_\pi(t, \mathcal{V}^{\mathcal{O}}) \leq f_{\pi_i}(t, \mathcal{V}^{\mathcal{O}}) \; \forall \pi_i \in \Pi$$
$$\mathcal{V}^{\mathcal{O}}[i] \leq V_{\max}, \; 1 \leq i \leq d$$

Note that the free variables in the system are the components of $\mathcal{V}^{\mathcal{O}}$. The objective function maximizes the Bellman error at state $s$ under the assumption that action $a$ is taken. The first set of constraints identifies the region in $\mathcal{V}^{\mathcal{O}}$ space for which $\pi$ dominates at $t$. If this region exists, it is guaranteed to be a single, continuous facet of a convex surface by Lemma 1. The last set of constraints bounds $\mathcal{V}^{\mathcal{O}}$ to be within the range of possible values.

The largest value returned by the linear program over all $s$, $a$, and $\pi$ provides the point at which the current cache of policies will have the largest Bellman error. The time bound is satisfied because linear programming is polynomial in the size of its inputs.■

This provides a basis for determining if a policy cache is $\epsilon$−optimal. If the largest Bellman error returned by the linear program in the above theorem is less than $\epsilon$, this means that no matter what values the states in the out-space of $G$ assume, the policy assigned to $G$ will produce state values with a Bellman error of less than $\epsilon$. Thus, if the largest Bellman error is less than $\epsilon$, the cache is $\epsilon$-optimal. The computational consequences of Theorem 2 are non-trivial and deserve emphasis: When combined with Theorem 1, this appears to be the first efficient method known for determining if a set of policies for a region of an MDP is sufficient to produce a global solution that is within a bound of the optimal global solution. Note that the conditions checked by Theorem 2 are sufficient, but have not been shown to be necessary; more efficient methods may be possible.

Assuming, that the minimum possible state value is $V_{\min}$, this theorem provides a means of constructing an $\epsilon$−optimal cache of policies. Suppose that Theorem 2 is implemented as a function, find-worst, that takes a policy cache and returns two values, the point at which the Bellman error is worst, and the magnitude of this this error.

```
Π = {optimal policy for V^O = (V_min,...,V_min)}
quit = false
Repeat until quit
    (worst-error, worst-point) = find-worst(Π)
    if worst-error > ε
        Π = Π ∪ {optimal policy for worst-point}
    else
        quit = true
```

This algorithm keeps adding policies to the policy cache until the policy cache is proven to be $\epsilon$−optimal. Each policy that is added covers at least one case where the current cache is inadequate. Note that, worst-point will always be at a corner of the convex facet defined by some dominating policy. In practice, it is preferable to add a policy that is optimal for a point slightly towards the interior of the facet instead of at a corner. Note also that this algorithm implicitly assumes that the each new policy that is added to the cache will improve the value of every state in the region when $\mathcal{V}^{\mathcal{O}} =$ worst-point. This will be true if actions are sufficiently noisy such that reducing the Bellman error in any state will produce at least a minute improvement in the value of other states. It is possible to construct models where this assumption does not hold and in such cases additional constraints must be added to the linear program to break ties between policies. (See Parr (1998) for a more detailed discussion of these points.)

This approach has some similarities to and was inspired by algorithms for partially observable MDPs (POMDPs) (see Lovejoy (1991) for a survey), and in particular, the Witness algorithm (Cassandra, Kaelbling, & Littman, 1994). The treatment of policies as linear functions, the maximum over which forms a convex surface, is common in the POMDP literature. The value space search algorithm uses a similar approach to that of the Witness to search a continuous space to find the place where the error in the current set of policies is largest. The Witness algorithm is a synchronous value iteration algorithm that searches through belief space for a partially observable problem. The value space search algorithm searches through the space of state values to find the point at which the error in a set of infinite horizon MDP value functions is the largest.

The value space search algorithm was used to find an $\epsilon$−optimal policy cache for room 1 of Figure 1. This sub-problem contains 25 states and has a fan-out of 2. Possible actions are right, left, up and down, but these actions are unreliable, resulting in movement in one of the three other axis-parallel directions 20% of the time. The discount factor used was 0.95. There are $4^{25} \approx 10^{14}$ possible policies for this subproblem. Of course, many of these are unreasonable policies that, for example, move the agent in circles. However, a variety of policies can still be induced by different values of the out-space states, even in such a simple problem. One would think that the agent would simply aim for the exit with the highest state value. However, the noise in the action model ensures that there is always some chance that the agent will wind up unintentionally exiting the wrong way. Thus, as the relative difference between the two out-space states increases, the optimal policy will take a more circuitous route towards the desired exit, hugging the walls to avoid accidentally getting too close to the undesired means of egress.

If the values of the states are assumed to be on $[0 \ldots 20]$, then the $\epsilon$-grid approach for this problem would require 4 *million* policies for $\epsilon = 0.01$. The value space search algorithm produced a policy cache with the same optimality guarantees with just 22 policies. For $\epsilon = 0.001$, the $\epsilon$-grid approach would require 400 *million* policies, while the value space search algorithm produced the same 22 policies.

In this particular case, the value space search algorithm has captured the intuition that this type of subproblem should not be that hard. A few seconds of computation has produced a small cache of that will ensure a nearly optimal solution for this region no matter what happens in any connecting region. This small subproblem is now decoupled and completely solved — at least for $\epsilon \geq 0.001$ and for problems where the neighboring states can assume values on $[0 \ldots 20]$. Any MDP satisfying these conditions and with an optimality requirement of no more than $\frac{0.001}{1-0.95} = 0.02$ will *never* need another policy defined on this region.

An important unanswered question is whether this algorithm is guaranteed to find a polynomial size $\epsilon$−optimal cache of policies if such a cache exists. The idea of creating a new policy near the point where the current policy cache performs the worst is plausible, but there is not yet a proof that this constructs a cache of policies that is in any way minimal. A drawback of this algorithm is that it constructs a large number of linear programs. This can be onerous if the number of states in the region is large. Of course, this price is paid only once, and the cache can be reused indefinitely in any MDP that contains the same subproblem. Moreover, many implementation tricks can be used to reduce the size and number of linear programs constructed. For example, the maximum Bellman error for any (entry-point,state,action,policy) quadruple is non-increasing as the policy-cache grows, so the solutions to previous linear programs can be cached across calls to find-worst. A new linear program is needed only if the cached error is greater the maximum error detected so far in in the current call to find-worst.

## 4.3  The Convex Hull Bounding Approach

This section sketches a third algorithm with a computational geometry flavor. This algorithm also provides slightly different optimality guarantees than the previous algorithms. It guarantees that a cache of policies will be $\epsilon$-optimal *at the high-level*. Recall that high-level actions correspond to policies at the low level. If a cache is $\epsilon$−optimal at the high level, this means that there is no low-level policy that could improve the value of a high-level state by more than $\epsilon$. In the four-room example, this would mean that no policy could improve the value of one of the connecting states by more than $\epsilon$. However, there could be a policy that could improve the value of some other state, for example, the state in the top left corner of the model, by more than $\epsilon$, but the expected effect of this change on any of the high-level states must be less than this.

High-level $\epsilon$-optimality implies that any policy starting from a high-level state (e.g. one of the states connecting the rooms) will have expected value within $\frac{\epsilon}{1-\beta}$ of optimal. This could be a problem, however, if the agent typically starts in some state that is not a high-level state. In such cases, the starting position of the agent can be treated as if it were a connecting state by adding it to the in-space of the enclosing region and constructing a policy cache as if it were a connecting state. If desired, every state could be treated as if it were an in-space state, ensuring full low-level optimality as well.

The algorithm presented in this section has run time that is exponential in $d$, the fan-out of the region, but unlike the $\epsilon$-grid approach, it does not depend explicitly on $1/\epsilon$ and unlike the value space search algorithm, it can avoid considering every state inside of a region if high-level $\epsilon$-optimality is sufficient. The algorithm relies upon the following formal results:

**Lemma 2** *For any point $\mathcal{V}^{\mathcal{O}}$, set of points, $\mathcal{V}_1 \ldots \mathcal{V}_{d+1}$, with set of policies, $\pi_1 \ldots \pi_{d+1}$, such that $\pi_i$ is optimal with respect to $\mathcal{V}_i$ and such that the $\mathcal{V}_i$ form a convex hull around $\mathcal{V}^{\mathcal{O}}$, the optimal policy with respect to $\mathcal{V}^{\mathcal{O}}$ at any state $s$ is bounded from below by $\max_i f_{\pi_i}(s, \mathcal{V}^{\mathcal{O}})$ and from above by the hyperplane containing each of the $(\mathcal{V}_i, f_{\pi_i}(s, \mathcal{V}_i))$.*

**Proof:** Bounding from below is obvious and follows from Lemma 1: the optimal policy at any point must do at least as well as the dominating policy in the cache. The bound from above is somewhat more subtle: Let $H_1$ be the hyperplane containing the $(\mathcal{V}_i, f_{\pi_i}(s, \mathcal{V}_i))$. Suppose that there exists some $\pi$ and corresponding $f_\pi$ such that for some $s$, $f_\pi(s, \mathcal{V}^{\mathcal{O}})$ is above $H_1$. Let $H_2$ be the hyperplane corresponding to the linear value function of this policy at $s$. There must exist some corner of the convex hull used to create $H_1$ (some $(\mathcal{V}_i, f_{\pi_i}(s, \mathcal{V}_i))$) where $H_2$ is above $H_1$, i.e., $f_\pi(s, \mathcal{V}_i) > f_{\pi_i}(s, \mathcal{V}_i)$. However, $f_{\pi_i}$ is known to be optimal with respect to $\mathcal{V}_i$, so this is a contradiction.∎

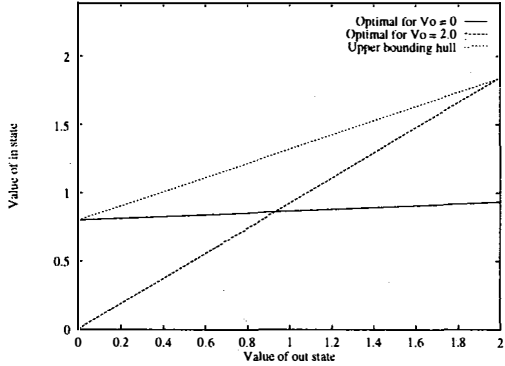

Figure 4: Two policies, and an upper surface bounding the their distance from the optimality.

A simple example of this lemma is shown in Figure 4. Two functions for policies from Figure 3 are shown. One is optima at $\mathcal{V}^\mathcal{O} = 0$ and the other is optimal at $V^\mathcal{O} = 2$. For any $0 \le \mathcal{V}^\mathcal{O} \le 2$, the linear function for the optimal policy cannot cross 0.8 at $\mathcal{V}^\mathcal{O} = 0$ or $\mathcal{V}^\mathcal{O} = 1.84$. Thus, the value of optimal policy is bounded by the line shown.

**Theorem 3** *For region $G$ and cache of policies, $\pi_1 \ldots \pi_m$, that are optimal at $\mathcal{V}_1 \ldots \mathcal{V}_m$, the optimal policy value for any $s$ with respect to any $\mathcal{V}^\mathcal{O}$ is bounded from below by the convex surface formed by the maximum over the corresponding $f_{\pi_1} \ldots f_{\pi_m}$ and bounded from above by the convex hull containing the points: $(\mathcal{V}_1, f_{\pi_1}(s, \mathcal{V}_1)), \ldots (\mathcal{V}_m, f_{\pi_m}(s, \mathcal{V}_m))$.*

**Proof:** The bound from below is a direct consequence of Lemma 1. The bound from above follows from Lemma 2 and noting that the lowest bounding hyperplane for any $\mathcal{V}^\mathcal{O}$ must form a facet in the convex hull of $(\mathcal{V}_1, f_{\pi_1}(s, \mathcal{V}_1)), \ldots (\mathcal{V}_m, f_{\pi_m}(s, \mathcal{V}_m))$.

This theorem also suggests an algorithm for finding points in value space where the current policy cache is not $\epsilon$−optimal: For each facet in the upper bounding hull, find the point in the lower hull that maximizes the distance between the two surfaces. If no point can be found where the distance is greater than $\epsilon$, then the policy cache is $\epsilon$ optimal. If the facets in the upper hull are enumerated as a set of linear functions, $g_1 \ldots g_l$, then the maximum distance can be checked for each $s$, $f_{\pi_i}$, and $g_j$ as follows:

*Maximize*

$$g_j(s, \mathcal{V}^\mathcal{O}) - f_{\pi_i}(s, \mathcal{V}^\mathcal{O})$$

Subject to

$$g_j(s, \mathcal{V}^\mathcal{O}) \le g_k(s, \mathcal{V}^\mathcal{O}) \ \forall k$$
$$f_{\pi_k}(s, \mathcal{V}^\mathcal{O}) \le f_{\pi i}(s, \mathcal{V}^\mathcal{O}) \ \forall k$$

$$\mathcal{V}^\mathcal{O}[k] \le V_{\max} \ \forall k$$

The last constraint bounds $\mathcal{V}^\mathcal{O}$ to lie in the permitted range. The first two sets of constraints identify the area in $\mathcal{V}^\mathcal{O}$ space in which $f_{\pi_i}$ is a facet on the lower bounding hull *and* $g_j$ is a facet on the upper bounding hull. If such an area exists, the objective function finds the point at which the distance from the upper hull to the lower hull is greatest.

The above linear program can be used to generate a cache of policies in a fashion similar to the value space search algorithm. By searching all pairs of upper bounding facets and lower bounding facets, the point at which the gap between these surfaces is greatest can be used to determine a new policy. One complicating factor is that the upper-bounding hull may not cover the entire space of values for $\mathcal{V}^\mathcal{O}$. For points outside the hull, the value of the optimal policy can be bounded from above by $V^{\max}$. In some cases this can be tightened by observing that no policy will do better than the sum of value of the optimal policy at $(V^{\min}, \ldots, V^{\min})$, and $\beta \max_i \mathcal{V}^\mathcal{O}[i]$, since the optimal policy for the lowest values of the out-space states will maximize the reward received within the region and no policy can do better than receiving this reward and then moving to the highest valued state in one step.

The more serious complication for this algorithm is the general result from computational geometry that the convex hull of $m$ points in $d$ dimensional space can have $O(m^{\lfloor \frac{d}{2} \rfloor})$ facets, making this algorithm exponential in $d$. Still, the convex hull bounding algorithm is superior to the $\epsilon$−grid approach since the $\epsilon$−grid approach has run time that depends directly on $\frac{1}{\epsilon}$ and the range of values possible in the out-space, while the bounding approach depends on the number of policies in the cache.

## 5    Partial Decoupling

The previous section presented 3 algorithms for completely decoupling MDPs. These algorithms are quite computationally intensive and there are no guarantees that the size of the policy cache required for a desired solution quality will be manageably small. In such cases, one may be forced to use a policy cache that is not known *a priori* to be $\epsilon$−optimal for the range of values the out-space of a region will take on when reconnected to the rest of the MDP. For example, a rough policy cache could be constructed for each of the rooms in a building. When a high-level problem that combines these rooms is solved, some decisions will need to be made on-the-fly about whether the policies in the rough cache are adequate for the larger problem.

More specifically, suppose an MDP has been divided into disjoint regions and a policy cache has been constructed for each region. A high-level decision problem can be defined over the out-spaces of these regions. For a particular region, $G$, an algorithm solving this high-level decision problem has the option of using one of the policies in the policy cache for $G$, or generating a new policy that is optimal for the algorithm's current estimate of $\mathcal{V}_G^{\mathcal{O}}$. A straightforward way to answer this question would be to use the cache $f_\pi$ functions to assign values to every state in the problem and then compute the Bellman error for each state. However, this approach would require so much computation that it would essentially defeat the purpose of solving a high-level problem. Instead, high-level optimality can be checked quite efficiently by using the tools of the convex hull bounding algorithm.

Starting with some policy cache, $\pi_1 \ldots \pi_m$, the elements of which are optimal at the corresponding $\mathcal{V}_1 \ldots \mathcal{V}_m$, for any particular $\mathcal{V}_G^{\mathcal{O}}$, the value of any state under the optimal policy with respect to $\mathcal{V}_G^{\mathcal{O}}$ is bounded from below by $\max_{\pi_i} f_{\pi_i}(s, \mathcal{V}^{\mathcal{O}})$, and the value is bounded from above the convex hull formed by $f_{\pi_1}(s, \mathcal{V}_1) \ldots f_{\pi_m}(s, \mathcal{V}_m)$ (Theorem 3). The situation here is slightly different from the bounding algorithm in that $\mathcal{V}_G^{\mathcal{O}}$ is fixed and known. Instead of a high-dimensional convex hull problem, the bounds for a *particular* $V^{\mathcal{O}}$ can be determined by solving a linear program. In the following $f_*$ is an unknown linear equation, i.e., the coefficients and constant are free variables:

*Maximize:*
$$f_*(s, \mathcal{V}^{\mathcal{O}})$$

Subject to:
$$f_*(s, \mathcal{V}_i) \leq f_{\pi_i}(s, \mathcal{V}_i), 1 \leq i \leq m$$
$$f_*(s, \mathcal{V}_{\mathcal{O}}) \leq V_{max}$$

To reassure oneself that this is indeed a *linear* program, recall that in this context, $\mathcal{V}^{\mathcal{O}}$, the $\mathcal{V}_i$, and coefficients and constants for the $f_{\pi_i}$ are all known constants. The only variables are the components of $f_*$. The first set of constraints requires that $f_*$ be no better than the optimal policy for $s$ at points in value space where the optimal policy is known. This is, essentially, a restatement of Lemma 2. The second set of constraints requires that $f_*$ never exceeds the maximum value any state can assume in this problem. Thus, the objective function forces the linear program to find the highest hyperplane that does not violate Lemma 2 or the bound on state values. If $\mathcal{V}^{\mathcal{O}}$ lies in the convex hull of $\mathcal{V}_1 \ldots \mathcal{V}_m$, then $f_*$ will be the facet of the upper-bounding convex hull from Theorem 3. Note that if $\mathcal{V}^{\mathcal{O}}$ does not lie in the convex hull, $V_{max}$ will be returned. This bound can be tightened by requiring that the constant of $f_*$ be no longer than the value of the optimal policy at $\mathcal{V}^{\mathcal{O}} = (V_{min} \ldots v_{min})$ and that the coefficients of $f_*$ sum to be no more than 1.

If the distance between the dominating policy and the upper bound returned by the above linear program is less than $\epsilon$ for every state in the in-space of $G$, then the policy cache for $G$ is sufficient to produce a high-level $\epsilon$-optimal policy for the current value of $\mathcal{V}^{\mathcal{O}}$. This means that a high-level decision problem can, for now, avoid updating the policy for region $G$ and focus attention on other regions. This decision will need to be reevaluated as values of the states in the out-space of $G$ change. One way to view this result is that it enables a form of high-level prioritized sweeping (Moore & Atkeson, 1993; Andre, Friedman, & Parr, 1998).

This result also has significant consequences for the transfer of knowledge across problems. Suppose, for example, that a particular model substructure appears in many different problems. Consider a larger version of the four-room problem with many interconnected rooms. Different tasks in this domain would correspond to different positions of the reward in different rooms. Every time a policy is produced for a room it can be added to the room's policy cache. The above linear program can be used to determine quickly if for some new problem, the cache in a particular room is adequate. Thus, a form of cross-task learning is achieved where the time required to plan for new objectives declines as experience is gained with the environment. Moreover, intelligent allocation of computational resources will be possible since parts of the value space that have already been mastered will no longer drain computational resources.

## 6    Conclusion

This paper presented two approaches to decoupling MDPs, a complete decoupling approach and a partial decoupling approach. With complete decoupling, the problem is divided into independent subproblems, and the solutions to these subproblems are combined in a single step. Two new algorithms for determining $\epsilon$-optimal policy caches for a subproblem are presented. The significance of the first algorithm is that it runs in polynomial time, regardless of the fan-out of the region. The second algorithm uses a computational geometry approach that can be exponential in the fan-out of the subproblem, but can be more efficient than the first algorithm if the fan-out is small.

Since complete decoupling may not always be possible, a method for partial decoupling is presented. This method assumes that an imperfect policy cache is used by a high-level asynchronous MDP algorithm. It uses the policy cache to bound the optimal values of states in a region with respect to the values of the states in the out-space of the region. By providing upper and lower bounds, this permits intelligent decisions about when to update the policy cache for a region based upon the algorithm's current estimate of the values of the states in the out-space of the region.

Together these results provide a framework for large-scale parallelization of MDPs and a formal framework for the transfer of knowledge across problems that share common structures. These results can be applied hierarchically, although the optimality requirements for the subproblems will become stricter with each division if the same level of optimality is to be maintained at the top level.

This work does not address the questions of state abstraction or value function approximation. Fortunately, these techniques will compliment the results presented here. The decoupled MDP algorithms will benefit from any approach that compresses the state space, especially if the compression reduces the fan-out of the regions in some decomposition of the space.

A limitation of this work is that it applies mainly to a restricted class of MDPs, those that are weakly coupled. Moreover, the efficiency of the methods described here will depend heavily upon the manner in which the MDP is decomposed into subproblems, and, in particular, the fan-out of the regions in the decomposition. The reader should keep in mind, however, that this appears to be one of the first attempts to aggressively decouple MDPs and that while the algorithms involved are, admittedly, complex, the potential benefits in parallelization and knowledge transfer across problems resulting from this line of research are substantial.

# 7    Acknowledgment

This work was support in part by DARPA contract DACA76-93-C-0025 under subcontract to Information Extraction and Transport, Inc., and through the generosity of the Powell Foundation and the Sloan Foundation and by DARPA Prime contract IET-1004-96-009. Some of this was done at the University of California at Berkeley, where it was supported in part, by ONR grant N00014-97-1-0942 and ARO MURI grant DAAH04-96-1-0341. The author benefited from helpful discussions about this and related work with David Andre, Craig Boutilier, Mike Bowling, Tom Dean, Nir Friedman, Milos Hauskrecht, Daphne Koller, Uri Lerner, Stuart Russell, Mehran Sahami and Rich Sutton. The reviewers also provided some extremely helpful comments.